\documentclass[10pt,journal,compsoc]{IEEEtran}
\ifCLASSOPTIONcompsoc
  \usepackage[nocompress]{cite}
\else
  \usepackage{cite}
\fi
\ifCLASSINFOpdf
\else
\fi
\usepackage{times}
\usepackage{color}

\usepackage{epsfig}
\usepackage{graphicx}
\usepackage{amsmath}
\usepackage{amssymb}
\usepackage{subfigure}
\usepackage{algorithm} 
\usepackage{algorithmic} 
\usepackage{amsmath}
\usepackage{color}
\usepackage{multirow}
\usepackage{bm}
\usepackage{amstext}
\usepackage{booktabs}
\usepackage{rotating}
\usepackage{balance}

\begin{document}
%
\title{Pairwise Instance Relation Augmentation for Long-tailed Multi-label Text Classification}
%
%
%
%

\author{Lin Xiao,
        Pengyu Xu, Liping Jing 
        and Xiangliang Zhang
\IEEEcompsocitemizethanks{
\IEEEcompsocthanksitem This work was supported in part by the National Key Research and Development Program of China under Grant 2020AAA0106800; The National Science Foundation of China under Grant 61822601 and 61773050; The Beijing Natural Science Foundation under Grant Z180006; The Fundamental Research Funds for the Central Universities (2019JBZ110).
\IEEEcompsocthanksitem Lin Xiao, Pengyu Xu and Liping Jing are with the Beijing Key Lab of Traffic Data Analysis and Mining, Beijing Jiaotong University, Beijing,
China, 100044.
\IEEEcompsocthanksitem Xiangliang Zhang is with the deparment of Computer Science and  Engineering, University of Notre Dame, USA.
\IEEEcompsocthanksitem Liping Jing is the corresponding author. Email: lpjing@bjtu.edu.cn.}
\thanks{Manuscript received Dec, 2021; revised Mar, 2022.}}

%
%

\markboth{IEEE Trans. on Knowledge and Data Engineering,~Vol.~**, No.~**, ****~2021}%
{Xiao \MakeLowercase{\textit{et al.}}: PIRAN}

%



\IEEEtitleabstractindextext{%
\begin{abstract}
Multi-label text classification (MLTC) is one of the key tasks in natural language processing. It aims to assign multiple target labels to one document. Due to the uneven popularity of labels, the number of documents per label follows a long-tailed distribution in most cases. It is much more challenging to learn classifiers for data-scarce tail labels than for data-rich head labels. The main reason is that head labels usually have sufficient information, e.g.,  a large intra-class diversity, while tail labels do not. In response, we propose a Pairwise Instance Relation Augmentation Network (PIRAN) to augment tailed-label documents for balancing tail labels and head labels. PIRAN consists of a relation collector and an instance generator. The former aims to extract the document pairwise relations from head labels. Taking these relations as perturbations, the latter tries to generate new document instances in high-level feature space around the limited given tailed-label instances. Meanwhile, two regularizers (diversity and consistency) are designed to constrain the generation process. The consistency-regularizer encourages the variance of tail labels to be close to head labels and further balances the whole datasets. And diversity-regularizer makes sure the generated instances have diversity and avoids generating redundant instances. Extensive experimental results on three benchmark datasets demonstrate that PIRAN consistently outperforms the SOTA methods, and dramatically improves the performance of tail labels.
\end{abstract}

}

\maketitle

\IEEEdisplaynontitleabstractindextext

%
\IEEEpeerreviewmaketitle

\section{Introduction}\label{sec:introduction}
Compared with single label text classification,  multi-label text classification (MLTC) is more complicated as it assigns multiple labels to one document.  Strong co-occurrences and inter-dependencies exist among multiple labels. For example, an article on the topic of religion is likely to talk about ethic as well, but unlikely to talk about weather. MLTC has been widely applied to diverse tasks from automatic annotation for topic recognition \cite{yang2016hierarchical,kazawa2004maximal,ueda2002parametric} to bioinformatics \cite{zhang2006multilabel,elisseeff2002kernel}, web mining \cite{tang2009large}, question answering~\cite{kumar2016ask}, tag recommendation \cite{katakis2008multilabel} and etc.

\begin{figure}[t]
	\centering
	\includegraphics[width=8cm]{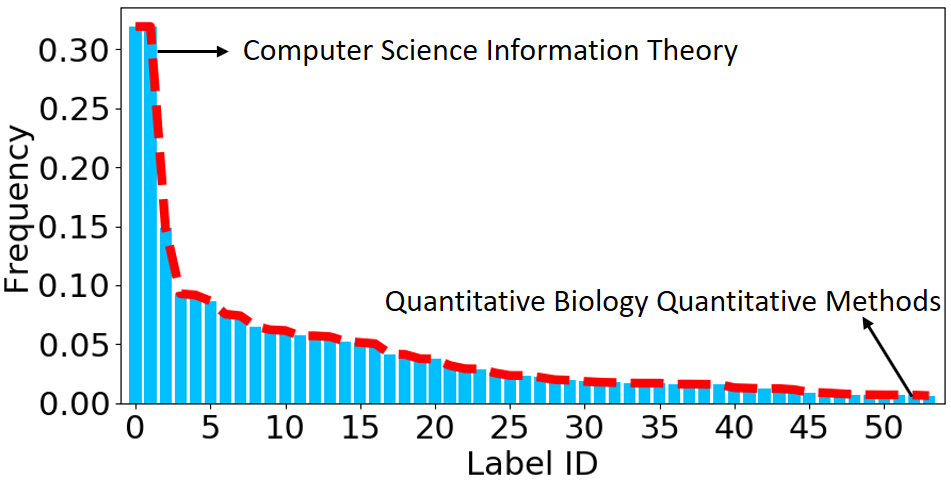}
	\label{Fig:model}
\caption{ The label frequency in AAPD dataset demonstrates a long-tail distribution.}
\label{Fig:label-distribution}
\end{figure}

Long-tailed distribution of the number of documents per label is a general issue in MLTC. Figure~\ref{Fig:label-distribution} shows the number of papers in the different fields of computer science on AAPD dataset collected from arXiv website. As shown in figure~\ref{Fig:label-distribution}, the long tail categories with very few training instances, e.g., the label named ``Quantitative Biology Quantitative Methods". It results the MLTC model performing exceptionally well on head labels (categories with a large number of training instances, e.g., the label named ``Computer Sciences Information Theory"), and the high-risk of over-fitting and inferior performance on tail labels separates hyperplane heavily skewed to the tail labels. 

Label correlations may provide helpful extra information, such information can help the classification of tail labels. Thus, the exploitation of label correlations has been widely accepted as a key solution of current multi-label learning approaches. To exploit label correlations, some approaches explore external knowledge such as existing label hierarchies \cite{cai2004hierarchical,cesa2006hierarchical,chen2020hyperbolic} or label semantic information\cite{xiao2019label,du2019explicit,rios2018few}. The co-occurrence of labels in training data is utilized in \cite{kurata2016improved,petterson2011submodular}. Although these approaches exploit label correlations in different ways to help alleviate long-tailed problem. However, they all train head labels and tailed labels together, as we know, it will make the head labels dominate the learning procedure, which will sacrifice the prediction precision of head labels and recall of tail labels. 

Many studies solve such long-tailed problem by artificially balancing strategies, such as data re-sampling strategies \cite{2003C4,chawla2002smote,buda2018systematic,byrd2019effect} and class-sensitive loss function \cite{2007The,2009Learning,2019Learning,lin2017focal,cui2019class,tan2020equalization}, which are designed to assign equal learning opportunities for each label.  Although such balancing strategies can alleviate the long-tailed problem to some extent, it may bring the following problems: (1) instance re-sampling tends to over-fit by duplicating tail data and damage feature representation when abandoning head data; (2) loss re-weighting may cause unstable training especially when the label distribution is extremely imbalanced. In other words, both strategies over-fitting on tail labels is still a big issue because of the limited intra-class diversity.

Transfer learning, as an alternative strategy, can promote the classification performance of the tail labels by transferring knowledge from head labels. Recently, several approaches based on transfer learning were proposed for long-tailed MLTC. 
LEAP \cite{liu2020deep} transfers the intra-class variance, OLTR \cite{liu2019large} transfers semantic deep features and HTTN \cite{xiao2021does} transfers the meta-knowledge between classifier parameters and label prototype. What they have in common is they learn rules or distributions of parameters from head labels, then transfer such meta-knowledge to tail labels. However, they cannot deliver adaptive adjustment to tail labels, i.e., transferring adapted meta-knowledge suitable for tail labels with few-shot instances.
To address this limitation, in this paper, we generate new instances for tail labels based on ``localization'' to alleviate the few-shot challenge. ``Localization'' means that we generate the new instances for tail labels by adding the pairwise instance relations on their prototypes, pairwise instance relations preserve instances neighbor structure of head labels which are expected to enlarge the intra-class diversity of tail labels. The classifiers of tail labels are then built in a many-shot situation. Compared to the previous transfer-based methods, the proposed  relation transfer explicitly complements what the tail labels need.

\begin{figure}[t]
	\centering
	\includegraphics[width=4cm]{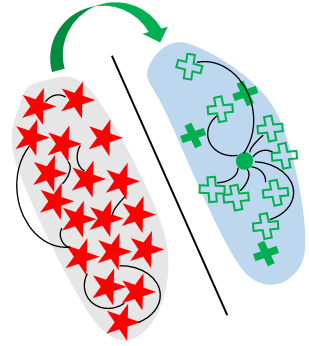}
	   \setlength{\abovecaptionskip}{0.cm}
	\setlength{\belowcaptionskip}{-0.cm}
\caption{Data augmentation by instance correlation transfer from a head label (with red star instances) to a tail label (with green cross instances).}
\label{Fig:correlation}
\end{figure}

The model is named Pairwise Instance Relation Augmentation Network (PIRAN), which has two main modules. The first module is a Relation Collector, which is implemented with the aid of head labels to get adequate instance relationships. The second module is  called Tail Instance Generator, which  transfers the relations to tail labels and generates new instances for tail labels in the feature space. As demonstrated in Figure~\ref{Fig:correlation}, we transfer the instance relations from head label to tail label, and then aggregate them with the prototype of tail label (with the green circle) to generate new instances for tail labels. The generated instances enrich the tail label distribution. 

To ensure the quality of the generated instances, we have to be aware of several issues. On the one hand, the variance of the new generated data may be too different from the variance of head labels. The transferred relations may drag the tail distribution with a wide spread. On the other hand, the generated data may be too similar when the transferred relations cause little perturbation to the feature space governed by a tail label. We thus constrain the new generations from two regularizers, consistency and diversity. For consistency, it contains two aspects, one is generation consistency, which ensures the new generations similar to the original tail prototype. Another is called variance consistency, which helps the variance of head and tail labels keep close. For the regularizer of diversity, it makes the generated representations be different, and effectively enlarges the intra-class diversity for tail labels. 

To make a summary, our main contributions are:
\begin{itemize}
\item A Pairwise Instance Relation Augmentation Network (PIRAN) is proposed to tackle the long-tailed problem in multi-label text classification. PIRAN takes advantage of the pairwise relations from the data-rich head labels to generate new instances for data-poor tail labels with the aid of Relation Collector and Instance Generator. 
\item To make the validity of new generations for tail labels, PIRAN designs two regularizers, consistency and diversity, which not only ensures the effectiveness of generation but also enlarges the intra-class diversity of tail labels.
\item The evaluation results on three widely-used benchmark datasets show that PIRAN outperforms the SOTA baselines,  especially on the prediction of tail labels.
\end{itemize}
The rest of the paper is organized via four Sections. Section 2 discusses the related work. Section 3 describes the proposed PIRAN model for multi-label text classification. The experimental setting and results are discussed in Section 4. A brief conclusion and future work are given in Section 5.

\section{Related Work}
In this section, we review previous literature from three aspects, multi-label text classification, long tail classification and long-tailed multi-label text classification.
\subsection{Multi-label Text Classification}
Multi-label classification is with wide applications in many areas, especially for text classification. Traditional methods can be grouped into two major categories: namely, problem transformation and algorithm adaptation. 
Problem transformation maps the MLTC task into multiple single-label learning tasks, such as BR~\cite{boutell2004learning}, LP~\cite{tsoumakas2007multi}, CC~\cite{read2011classifier}. Algorithm adaptation  extends specific multi-class algorithms to deal with multi-label classification problems
such as Rank-SVM ~\cite{elisseeff2002kernel}, ML-DT~\cite{clare2001knowledge}, ML-KNN~\cite{zhang2007ml}. Brieﬂy, the key idea of transformation methods is to ﬁt data to algorithm, while the key idea of algorithm adaptation methods is to ﬁt algorithm to data.

Recent years, with the development of neural network, it have become the promising solution for the multi-label classification. Among them, recurrent convolutional neural network (RCNN)~\cite{lai2015recurrent} captures contextual information with the recurrent structure and constructs the representation of text using a convolutional neural network. XML-CNN~\cite{liu2017deep} captures the local correlations from the consecutive context windows. Although obtains promising results, but it suffer from the limitation of window size so it cannot determine the long-distance dependency of text. Besides them, attention mechanism is introduced to yield unusually brilliant results on MLTC. SGM \cite{yang2018sgm} uses sequence generation model with attention structure to solve the MLTC task. AttentionXML~\cite{you2019attentionxml} proposes a multi-label attention to capture the most relevant parts of texts to each label and uses the label probabilistic label tree to tackle MLTC. However, most of them ignore the correlation among labels. An initialization method ~\cite{kurata2016improved} is proposed to treat some neurons in the final hidden layer as dedicated neurons for each pattern of label co-occurrence. Some methods are proposed to capture label correlations by exploiting label structure or label content. DXML~\cite{zhang2018deep} constructs the label graph from the label co-occurrence to supervise the MLTC. EXAM~\cite{du2019explicit} leverages the interaction mechanism to explicitly compute the word-level interaction signals for the text classification. Although multi-label learning benefits from the exploration of label correlation, these methods can not well handle long tail problem because most of them treat the labels equally. In this case, infrequently occurring tail labels are harder to predict than head labels since they have little training documents which make the whole learning model dominates by the head labels~\cite{jain2016extreme}.
\subsection{Long Tail Classification}
Data re-sampling~\cite{drumnond2003class,han2005borderline,2017A,chawla2002smote,byrd2019effect} is one traditional solution for long-tail learning. It typically over-samples the training data from tail labels, or equivalently to under-sample head labels. 
Another traditional solution is re-weighting, it modifies the loss function that is used to train the model~\cite{lin2017focal,cao2019learning,tan2020equalization,huang2016learning}. Their common idea is to assign a higher loss to samples from tail labels, thereby providing a stronger supervisory signal for the model to learn these labels. Recent studies shows a trend of decoupling long-tailed classification into two separate stages: representation learning and classifier learning~\cite{kang2019decoupling,zhou2020bbn}. BBN~\cite{zhou2020bbn} proposes a unified Bilateral-Branch Network (BBN) integrating “conventional learning branch” and “re-balancing branch”. The former branch is equipped with the typical uniform sampler to learn the universal patterns for recognition, the later branch is coupled with a reversed sampler to model the tail data. Similar idea is used in ~\cite{kang2019decoupling}. Both of them empirically prove that re-balancing methods may hurt feature learning to some extent. 
Transfer learning as an effective strategy, it transfer the knowledge learned from data-rich head classes with abundant training instances to help data-poor tail classes. \cite{wang2017learning} gradually transfer meta-knowledge learned from the head to the tail through a recursive approach. In~\cite{liu2019large},  an integrated algorithm is developed with dynamic meta-embedding. It handles tail recognition robustness by relating visual concepts among head and tail label embedding. LEAP~\cite{liu2020deep} expands the distribution of tail labels during training by transferring the intra-class angular distribution learned from head labels to tail labels. All the methods mentioned above are proposed to deal with long-tailed multi-class problems, the characteristic of multi-label is not fully considered.

\subsection{Long-tailed Multi-label Text Classification}
For MLTC task, the re-sampling strategy may not necessarily result in a more balanced distribution of classes due to the label correlations. For example, a document that contains tail labels, e.g. ``Condensed Matter Statistical Mechanics'' and ``Symbolic Computation'', is likely to be also associated with head labels, e.g. ``Computer Science''.  
To solve the multi-label long-tailed problem, some methods take advantage of label correlations and extra label information to alleviate the imbalance between head labels and tail labels in some extent, they still can not well handle long tail problem. To tackle the long-tailed multi-label problem, \cite{yuan2019long} proposes a linear ensemble by using DNN and linear classifier for head and tail labels respectively.  \cite{macavaney2020interaction} introduces the label semantics and then employ soft n-gram interaction matching to tackle tail labels. ZACNN \cite{rios2018few} introduces a neural architecture that incorporates label descriptors and the hierarchical structure of the label spaces for few and zero-shot multi-label text prediction.  HTTN \cite{xiao2021does} learns meta-knowledge for mapping prototypes learned by few-shot to those by many-shot. 

It is worth mentioning that above-discussed related work gain the knowledge mainly from the head labels but they don't consider whether the transferred knowledge is suitable for tail labels. Different from them, we transfer the pairwise correlations to generate the new instances for tail labels. And the generated instances adapt the correlation of each pair instances in head labels to tail labels. The classifiers of tail labels are then built in a many-shot situation. Comparing to the parameter-level knowledge transfer, the proposed data augmentation with correlation transfer explicitly complements what the tail labels need.
\begin{figure*}[htbp]
	\centering
	\includegraphics[width=15cm]{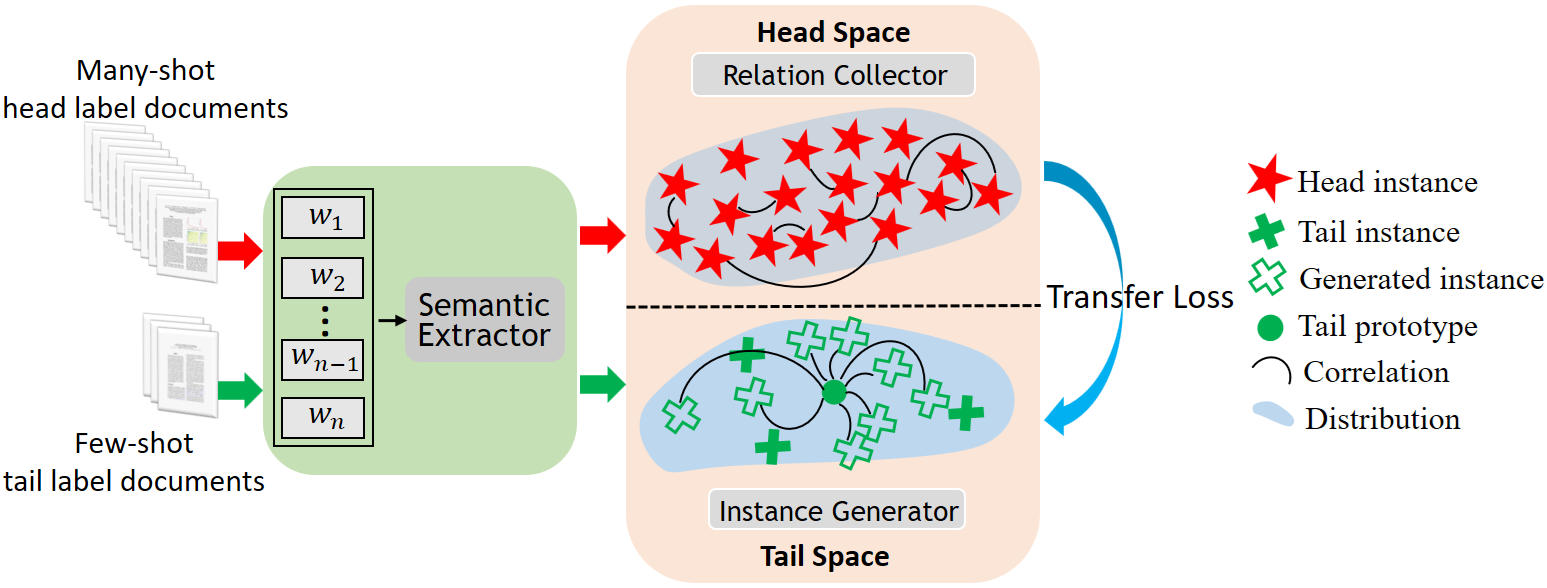}
\caption{Pairwise Instance Relation Augmentation Network (PIRAN). The Relation Collector gets adequate instance relationships in head labels. The Tail Instance Generator transfers the correlation to tail labels and generates the new instances for tail labels.  }
\label{Fig:framework}
\end{figure*}
\section{Proposed Method }
Let $D = \{(x_{i},y_{i})\}_{i=1}^{N}$ be a set of  $N$ documents with corresponding labels $Y=\{y_{i}\in \{0,1\}^{l}\}$, here $l$ is the total number of labels. Each document contains a sequence of words, $x_i=\{w_{1},\cdots,w_{j},\cdots,w_{n}\}$, where
$w_{j}\in \mathbb{R}^{k}$ is the $j$-th word vector (e.g., encoded by word2vec~\cite{pennington2014glove}), and $n$ is the number of words in the document.
Multi-label text classification (MLTC) aims to learn a classifier from $D$, which can assign the most relevant labels to the new given documents. In this study, we divide  the label set into two parts: \emph{head} labels that are associated with many documents (higher than a given threshold), and \emph{tail} labels that are associated with few documents (lower than a given threshold). The number of head labels and tail labels are $l_{head}$ and $l_{tail}$, respectively. 

In long-tailed multi-label text classification task, head labels have sufficient documents that can be used to train high-performance classifiers. On the contrary, there are only few-shot documents for tail labels to build their classifiers. The imbalanced distribution over labels often causes poor performance on tail labels when head and tail labels are treated with no difference. Here, we consider to design a data augmentation strategy to calibrate this biased distribution and alleviate the difficulties for few-shot tail label learning. 
A Pairwise Instance Relation Augmentation Network (PIRAN) is proposed to transfer the instance relations from head labels to tail labels for generating new instances to enlarge the intra-class diversity of tail labels. The overall architecture of PIRAN is shown in Figure~\ref{Fig:framework}. Besides the  Semantic Extractor for learning comprehensive document representation,
PIRAN has two other main modules, Relation Collector and Tail Instance Generator. Relation Collector captures the instance correlation from head labels, and Tail Instance Generator applies these correlation pairs to tail labels to generate new instances. 

\subsection{Document Representation by Semantic Extractor}
To capture the forward and backward sides contextual information of each word, we adopt the bidirectional long short-term memory (Bi-LSTM)~\cite{hochreiter1997long} language model to learn the word embedding for each input document. At time-step $p$, the hidden state can be updated with the aid of input and $(p-1)$-th step output.
\begin{equation}\label{1}
\begin{split}
\overrightarrow{h_{p}} = LSTM(\overrightarrow{h_{p-1}},w_{p})\\
\overleftarrow{h_{p}} = LSTM(\overleftarrow{h_{p-1}},w_{p})
\end{split}
\end{equation}
where $w_p$ is the embedding vector of the $p$-th word in the corresponding document, and $\overrightarrow{h_{p}},\overleftarrow{h_{p}}\in \mathbb{R}^{k}$ indicate the forward and backward word context representations respectively.

Then, the whole document can be represented by Bi-LSTM as follows.
\begin{equation}\label{hlstm}
\centering
\begin{split}
H &= (\overrightarrow{H} ,\overleftarrow{H}).  \\
\overrightarrow{H} &= (\overrightarrow{h_{1}},\overrightarrow{h_{2}},\cdots,\overrightarrow{h_{n}})\\
\overleftarrow{H} &= (\overleftarrow{h_{1}},\overleftarrow{h_{2}},\cdots,\overleftarrow{h_{n}})
\end{split}
\end{equation}
In this case, the whole document set can be taken as a matrix $H \in \mathbb{R}^{2k\times n}$.
As mentioned above, a multi-label document may be tagged by more than one label. To capture the components relevant to different labels in each document, we adopt the multi-head attention mechanism as the Semantic Extractor.
The multi-head document representation can be obtained by
\begin{equation}\label{as}
\begin{split}
&A = softmax(W_{2}tanh(W_{1}H))\\
&M_{\tau\cdot} = A_{\tau}H^{T}\\
&r = f_{a}(M)
\end{split}
\end{equation}
where $W_{1}\in \mathbb{R}^{d_{a}\times 2k}$ and $W_{2}\in \mathbb{R}^{d\times d_{a}}$
are so-called attention parameters to be trained. The dimension parameter $d_a$  can be set arbitrarily. The matrix $A\in \mathbb{R}^{s\times n}$ is the multi-head attention score for $n$ words in the document. Here,  $s$ is the number of heads in the multi-head attention mechanism, and  $\tau$ means the $\tau$-th head of attention. The multi-head document representation $M\in\mathbb{R}^{s\times 2k}$ is obtained by using linear combination of the context words with the aid of attention score ($A$). The comprehensive document representation $r \in \mathbb{R}^{d}$ is produced by aggregating the multi-head representation $M$ with an aggregation function $f_{a}$.
The Semantic Extractor is shared between head and tail labels, and it can be replaced with any modern language model such as BERT~\cite{devlin2018bert}, GPT-3~\cite{brown2020language} and etc.

Once having the comprehensive document representation, we can build the multi-label text classifier. Mathematically, the predicted probability of each label for a document can be estimated via
\begin{equation}\label{5}
\hat{y} = sigmoid(W_{a}r)
\end{equation}
where $W_{a}\in \mathbb{R}^{l\times d}$ is the classifier parameter for all labels, including both the classifier parameter of head labels $W_{head}$ and tail labels $W_{tail}$. 
The sigmoid function is used to transfer the output value into a probability.  The cross-entropy is often used for constructing the loss function  for multi-label text classification task~\cite{nam2014large}. 
\begin{equation}\label{eq:classifier-loss}
\begin{split}
\mathcal{L}=&-\sum^{N}_{i=1}\sum_{j=1}^{l}y_{ij}\log(\hat{y}_{ij})\\
&+(1-y_{ij})\log(1-\hat{y}_{ij})
\end{split}
\end{equation}
where $N$ is the number of training documents, $l$ is the number of labels, $\hat{y}_{ij}\in [0,1]$ is the predicted probability, and $y_{ij} \in\{ 0,1\} $ indicates the ground truth of the $i$-th document along the $j$-th label. To prevent the predicted scores from being biased toward frequent head labels, we re-train the classifier of tail labels according to data augmentation strategy with the aid of Relation Collector and Tail Instance Generator.
\begin{table*}[t]
\begin{center}\label{Tabel:dataset}
\caption{\label{font-table} Summary of Experimental Datasets. $N$ is the number of training instances, $M$ is the number of test instances, $D $ is the total number of words, $ L $ is the total number of classes, $\bar{L} $ is the average number of labels per document, $\tilde{L}$ is the average number of documents per label, $ \bar{W}$ is the average number of words per document in the training set, $ \tilde{W} $ is the average number of words per document in the testing set. }
\begin{tabular}{c|cccccccc}
\hline  Datasets &  $N$ &  $M$ &  $D$ &  $L$ &  $\bar{L}$ &  $\tilde{L}$ &  $\bar{W}$ &  $\tilde{W}$\\ \hline
RCV1 & 23,149  & 781,265 & 47,236 & 103 & 3.18& 729.67 & 259.47 & 269.23\\
AAPD & 54,840  & 1,000 & 69,399 & 54 & 2.41 & 2444.04  & 163.42 & 171.65\\
EUR-Lex & 13,905  & 3,865 &33246 & 3,714 & 5.32& 19.93  & 1,217.47 & 1,242.13\\
\hline
\end{tabular}
\end{center}
\end{table*}
\subsection{Relation Collector}
Instances belonging to head labels demonstrate abundant pairwise relations that are missing in tail labels. The pairwise relations preserve instances neighbor structure information. To generate instances for tail labels, we can copy these pairwise relations and adjust them in the tail cases. For example, the most intuitive operation is to transfer the pairwise relations to tail labels by scaling operations.
To get the transferable pairwise relations, Relation Collector is designed to make an efficient collection. 
Specifically, for each head label, we sample $p$ pairs of instances and measure the difference between each pair of them. The collected relations for a head label is $\{c_{1}, \cdots, c_{p} \}$, where one relation $c$ is calculated as follows 
\begin{equation}
\begin{split}
c=r_1-r_2\\
\end{split}
\end{equation}
$c$ is the difference between the two sampled instances $r_1$ and $r_2$ in one pair. 
From all head labels, we extract in total $l_{h}\times p$ relations to transfer for tail instance generation. It is analogous to allow some teachers (head labels) to impart knowledge (instance relations) to their students (tail labels) from what they have. 

\subsection{Instance Generator}
For a tail label $j$, we sample $q$ documents and get their representation $\{r^{j}_{1},\cdots, r^{j}_{q}\}$ by the trained multi-head Semantic Extractor. Note that $q$ can be a small value, e.g., 1 to 5. 
Then, we use the average function to get the prototype of the tail label,
\begin{equation}
\begin{split}
o^{j}& = avg\{r^{j}_{1},\cdots, r^{j}_{q}\}\\
\end{split}
\end{equation}
Although the multi-head semantic extractor can map the documents of tail labels to comprehensive representations, the obtained  prototype $o^{j}$ is not capable to govern the same region that could be reached at the condition that tail label $j$ has a sufficient number of instances. To make an instance-rich environment for label $j$, we can generate new instances centering at prototype $o^{j}$ by transferring the  collected relations from head labels. 

By using the $p$ relations from one head label $b$, we can generate $p$ instances for a label $j$, 
\begin{equation}
\begin{split}
g^{jb}_{z}& = W(o^{j} + c^{b}_{z}), \; z = [1,\cdots,p ]
\end{split}
\end{equation}
where $c^{b}_{z}$ is one of the $p$ relations from head label $b$, and $g^{jb}_{z}$ is one new generated instance. Trainable parameter $W$ works as scaling operation to make the perturbation $c^{b}_{z}$ on prototype $o^{j}$ be adapted to the tail label setting. 

The parameter $W$ is trained by constraining the new generations on two regularizers, consistency and diversity. They  are reflected in three loss functions defined below, generation and variance consistency, generation diversity.

\begin{equation}
\mathcal{L}_{gen}=\sum_{j=1}^{l_{t}}\sum_{b=1}^{l_{h}}\sum_{z=1}^{p}||o^{j}-g^{jb}_{z}||^2
\end{equation}

\begin{equation}
\begin{split}
&\mathcal{L}_{div}=-\sum_{j=1}^{l_{t}}\sum_{b=1}^{l_{h}}\sum_{z=1}^{p}||Q_{h}^{T}g^{jb}_{z}-\frac{1}{p}\sum_{z=1}^{p}Q_{h}^{T}g^{jb}_{z}||^2
\end{split}
\end{equation}

\begin{equation}
\mathcal{L}_{var}=\sum_{j=1}^{l_{t}}\sum_{b=1}^{l_{h}}\sum_{z=1}^{p}||Q_{h}Q_{h}^{T}g^{jb}_{z}-g^{jb}_{z}||^2
\end{equation}
The loss ${L}_{gen}$ is called generation consistency. It is designed to make the new generations for one tail label be close to its label prototype, so that the overall distribution of the generated instances is roughly located close to the few-shot distribution. However, if the new generations are too close,  the generated instances will be redundancy and invalid due to the low  intra-class diversity. Therefore, generation diversity ${L}_{div}$ is designed to combat this problem by measuring the projected variance of generated instances. The projected space is spanned by $Q$, which are the eigenvectors calculated from the head labels' covariance matrix:
\begin{equation}
\begin{split}
\sum_{b=1}^{l_{h}}\sum_{z=1}^{n_{l_{h}}}(r_{bz}-o^{b})^{T}(r_{bz}-o_{b})
\end{split}
\end{equation}
 where $o_{b}$ is the prototype of head label. The intra-class variance are evaluated from individual documents of head labels to tail labels. We take the top 100 eigenvectors as preserving 95\% information.
The variance consistency is introduced by ${L}_{var}$ that measures the reconstruction loss of generated instances in the space of $Q$. This  further ensures the generated instances for tail labels have a consistent variance w.r.t. the head labels.   
These three loss functions are combined with coefficients $\alpha=1$, $\beta=1$, $\gamma=0.1$ to be the instance correlation transfer loss. These coefficients were set by grid search to reach the best performance on training $W$.
\begin{equation}
\mathcal{L}_{transfer}=\alpha\mathcal{L}_{gen}+\beta\mathcal{L}_{var}+\gamma\mathcal{L}_{div}
\end{equation}
\subsection{Classifier Adjustment}
Once we have the new instances and their representations for tail labels, we use these representations to adjust the tail classifier parameter $W_{tail}$, such that the tail class prediction
\begin{equation}
\begin{split}
\hat{y} = sigmoid(W_{tail}g^{jb}_{z})
\end{split}
\end{equation}
\begin{algorithm}[t]
\caption{Pairwise Instance Relation Augmentation Network }
\renewcommand{\algorithmicrequire}{\textbf{Input:}}
\renewcommand{\algorithmicensure}{\textbf{Output:}}
\label{alg}
\begin{algorithmic}[1]
\REQUIRE Documents $D = \{(x_{i},y_{i})\}_{i=1}^{N}$
\ENSURE Prediction label set $\hat{y}$ for new coming document $x$.\\
Stage1:\\
\WHILE{ not converge }
    \STATE {use Semantic Extractor Eq.(1), (2), (3) to obtain the document representation;}
    \STATE {predict labels by using Eq.(4);}
    \STATE {update the Semantic Extractor and classifier parameters by evaluating loss function Eq.(5);}
\ENDWHILE
\\
Stage2:\\
\FOR  {$j$ in $l_{head}$}
\STATE {sample $p$ pair of documents with label $j$;}
\STATE {obtain the instance relations $\{c_{1}, \dots, c_{p}\}$ by using Relation Collector in Section 3.2;}
\ENDFOR
\WHILE{ not converge }
\FOR  {$j$ in $l_{tail}$}
\STATE {sample $q$ documents with label $j$;}
\STATE {obtain the prototype $o_{j}$ by using Eq.(6);}
\STATE {generate $p$ instances for tail label $j$ by use Eq.(7);}
\STATE{update the regularize parameter $W$ to keep the generation and variance consistency, generation diversit by use Eq.(11);} 
\ENDFOR
\ENDWHILE
\STATE {update the classifier parameters of tail labels $W_{tail}$ by using new generations and Eq.(12);}
\end{algorithmic}
\end{algorithm}
can minimize the loss function defined in Eq.(\ref{eq:classifier-loss}).  The adjusted tail classifier $W_{tail}$ is expected to work like have seen many-shot instances.
which is concatenated with the head label classifier, forming  the whole classifier for inference:
\begin{equation} 
W_{a}= cat[W_{head}:\hat{W}_{tail}]
\end{equation}

Algorithm 1 illustrates the procedure of PIRAN in detail. The lines 1-5 present the learning process of Semantic extractor and classifier parameters. The lines 6-9 show the procedure of Relation Collector which obtains the instance relations from head labels.
The lines 10-17 present the Instance Generator procedure of tail labels with the aid of instance relations, which obtains from head labels. The line 18 updates the  parameter of tail labels by using the new generations.

Given a testing document, it will first go through the Semantic Extractor to have its representation vector $r$, and then get the predicted label by $\hat{y}=sigmoid(W_{a}r)$. $ W_{a}$ is composed of $W_{head}$ and $W_{tail}$. $W_{tail}$ is the parameter of tail labels, which has been adjusted by many-shot instances.
\section{ Experiments}
In this section, we evaluate the proposed model PIRAN on three datasets by comparing with seven state-of-the-art methods in terms of widely used metrics, P@k, nDCG@k (k=1,3,5) and F1-score.
\subsection{Experimental Setting}
\textbf{Datasets.}
Three multi-label text datasets are employed for evaluation:  AAPD, RCV1 and EUR-Lex. Their label distributions all show the long-tailed distribution.
\begin{itemize}
    \item {\bf AAPD }~\cite{yang2018sgm} collects the abstract and the corresponding subjects of 55840 papers from arXiv in the filed of computer science.
    \item {\bf Reuters Corpus Volume I (RCV1)}~\cite{lewis2004rcv1} contains more than 80K manually categorized news belonging to 103 classes.
    \item{\bf EUR-Lex}~\cite{mencia} is a collection of documents about European Union law belonging to 3714 subjects (labels).
\end{itemize}
For these three data sets, only last 500 words were kept for each document. Once the document has less than the  predifined number of words, we extend it by padding zeros.
These widely used benchmark datasets have defined the training and testing split. We follow  the same  data usage for all evaluated models. The datasets are  summarized in Table \ref{font-table}.\\
\textbf{Baseline Models.} To verify the effectiveness of PIRAN, we selected the seven most representative baseline models in three groups.\\
\textbf{1) Multi-Label Methods:}\\
\vspace{-1.2em}
\begin{itemize}
\item {\bf DXML:}~\cite{zhang2018deep} tries to explore the label correlation by considering the label structure from the label co-occurrence graph. 
\item {\bf XML-CNN:}~\cite{liu2017deep} adopts Convolutional Neural Network (CNN) and a dynamic pooling technique to extract high-level feature for multi-label text classification. 
\item{\bf AttentionXML:}~\cite{you2019attentionxml} builds the label-aware document representation for multi-label text classification.
\end{itemize}
\textbf{2) Long-Tailed Multi-Class Methods:}\\
\vspace{-1.2em}
\begin{itemize}
\item {\bf OLTR}~\cite{liu2019large} learns dynamic meta-embedding in order to share visual knowledge between head and tail classes.
\item {\bf LEAP}~\cite{liu2020deep} transfers the intra-class diversity learned from head labels to tail labels.
\end{itemize}
\textbf{3) Long-Tailed Multi-Label Methods:}\\
\vspace{-1.2em}
\begin{itemize}
\item {\bf LTMCP}~\cite{yuan2019long} introduces an ensemble method to tackle long-tailed multi-label training. The DNN and linear classifier are combined to deal with the head label and tail label respectively. 
\item {\bf HTTN}~\cite{xiao2021does} captures meta-knowledge from head to tail for mapping few-shot network parameters to many-shot network parameter.
\end{itemize}
\begin{table*}[tp]
\begin{center}
\caption{\label{font-table1} Comparing PIRAN with baselines on AAPD dataset.}
\begin{tabular}{|c|ccc|cc|c|}
\hline
&P@1&P@3&P@5&nDCG@3&nDCG@5&F1-score\\
\hline
DXML&80.54&56.30&39.16&77.23&80.99&65.13\\
\hline
XML-CNN&76.25&54.34&37.84&72.01&76.40&65.52\\
\hline
AttentionXML&83.02&58.72& 40.56&78.01&82.32&68.79\\
\hline
OLTR&78.96&56.28&38.60&74.66&78.58&62.48\\
\hline
LEAP& 82.23 & 60.00& 40.95&79.36&83.21&68.09\\
\hline
LTMCP&78.51&56.02&38.46&75.19&76.05&63.59\\
\hline
HTTN&82.79&58.69&40.18&77.15&81.32&68.47\\
\hline
PIRAN & \bf{83.94} & \bf{60.61} & \bf{41.65} & \bf{79.57} & \bf{83.74}& \bf{69.81} \\
\hline
\end{tabular}
\end{center}
\begin{center}
\caption{\label{font-table2} Comparing PIRAN with baselines on RCV1 dataset.}
\begin{tabular}{|c|ccc|cc|c|}
\hline
&P@1&P@3&P@5&nDCG@3&nDCG@5&F1-score\\
\hline
DXML&94.04&78.65& 54.38&89.83& 90.21 &75.76\\
\hline
XML-CNN&95.75&78.63&\bf{54.94}&89.89&90.77&75.92\\
\hline
AttentionXML&95.62&{78.72}&54.22&90.21&90.59&77.95\\
\hline
OLTR&93.79 &61.36& 44.78 &74.37& 77.05& 56.44\\
\hline
LEAP& 95.64& 78.84&54.90&89.97&90.74&79.52\\
\hline
LTMCP&90.76 &73.26& 51.62&84.31&85.12&74.28\\
\hline
HTTN&94.00&75.79&53.87&86.90&87.65&76.27\\
\hline
PIRAN&\bf{95.87}& \bf{79.21}& 54.76& \bf{90.40}& \bf{91.09}& \bf{79.84}\\
\hline
\end{tabular}
\end{center}

\begin{center}
\caption{\label{font-table3} Comparing PIRAN with baselines on EUR-Lex dataset.}
\begin{tabular}{|c|ccc|cc|c|}
\hline
&P@1&P@3&P@5&nDCG@3&nDCG@5&F1-score\\
\hline
DXML&80.41 & 66.74&56.33 &70.03& 63.18& 53.28\\
\hline
XML-CNN&78.20& 65.93& 53.81 &68.41& 60.54 &51.98\\
\hline
AttentionXML&78.36&66.01&53.97&68.74&63.29&52.75\\
\hline
OLTR&65.62&52.34 &42.69 &55.73 &50.57& 22.64\\
\hline
LEAP&80.02&64.09&51.73&68.16&61.57&47.64\\
\hline
LTMCP&75.23& 60.12 &49.36 &64.89 &58.23& 48.10\\
\hline
HTTN&79.58&65.28&54.92&68.46&63.71&53.34\\
\hline
PIRAN&\bf{80.56}& \bf{67.37}& \bf{56.47}& \bf{70.99}& \bf{65.21}& \bf{54.68}\\
\hline
\end{tabular}
\end{center}
\end{table*}
\textbf{Parameter Setting}\\
For all three datasets, we use Glove~\cite{pennington2014glove} to get the word embedding in 300-dim.  LSTM hidden state dimension is set to $k=150$. The parameter $d_a=100$ for $W_1$ and $W_{2}$. $s=4$ for $M$ and $d=100$  for $r$ and $W_{all}$.  The whole model is trained via Adam~\cite{kingma2014adam} with the learning rate being 0.01. The baselines OLTR and LEAP deal with the long tail problem on the image recognition, the Semantic Extractor used was the ResNet-10, ResNet-32 and others. For a fair comparison, we replace the Semantic Extractor with Bi-LSTM with attention. All experiment results reported in this paper are averaged from five independent runs. The parameters of all baselines are either adopted from their original papers or determined by experiments.
For AAPD, RCV1 datasets, we set the number of tail labels as 24 (44.4\%) and 60 (60\%), respectively. For Eurlex dataset, we set the number of tail label be 2476 (66.7\%).\\
\textbf{Evaluation Metrics}. We use the metric, precision at top K ($P@k$) and the Normalized Discounted Cumulated Gains at top $K$ ($nDCG@k$), F1-score to evaluate the prediction performance. $P@k$ and $nDCG@k$ are defined according to the predicted score vector $\hat{y}\in \mathbb{R}^{l}$ and the ground truth label vector $y\in \{0,1\}^{l}$ as follows.
\begin{equation}\label{5}
\begin{split}
&P@k = \frac{1}{k}\sum_{l\in rank_{k}(\hat{y})}y_{l} \\
&DCG@k = \sum_{l\in rank_{k}(\hat{y})}\frac{y_{l}}{\log(l+1)}\\
 &nDCG@k= \frac{DCG@k}{\sum_{l=1}^{\min(k,\|y\|_{0})}\frac{1}{\log(l+1)}}
\end{split}
\end{equation}
where $rank_{k}(y)$ is the label indexes of the top $k$ highest scores of the current prediction result. $\|y\|_{0}$ counts the number of relevant labels in the ground truth label vector $y$. F1-score is the harmonic mean of Precision and Recall, Precision($P_{i}$), Recall($R_{i}$), and F1-score can be computed as below:
\begin{equation}\label{5}
\begin{split}
&F1-score=\frac{1}{|l|}\sum_{i\in l}\frac{2P_{i}}{R_{i}}{P_{i}+R_{i}}\\
&P_{i}=\frac{{TP}_{i}}{TP_{i}+FP_{i}}\\
&R_{i}=\frac{TP_{i}}{TP_{i}+FN_{i}}
\end{split}
\end{equation}
where $TP_{i}$ ,${TN_{i}}$, $FP_{i}$ , $FN_{i}$ denote the true positives, true negatives, false positives and false negatives for the category $i$ in label set $l$, respectively.
\subsection{Comparison Results and Discussion}
 Table \ref{font-table1}, Table \ref{font-table2}, and Table \ref{font-table3} show the averaged performance of all test documents. According to the formula of P$@$K and nDCG$@$K, we know P$@$1 = nDCG@1, thus only nDCG@3 and nDCG@5 are listed. The best result is marked in bold.
From Table \ref{font-table1}, Table \ref{font-table2}, and Table  \ref{font-table3}, we can make several observations about these results.
Firstly, XML-CNN is worse than DXML and AttentionXML, it proposes a dynamic max pooling
scheme that captures richer information from different regions of
the document and a hidden bottleneck layer for better representations of documents as well as for reducing model size, but it ignores the label correlation, which is in fact the key focus for multi-label learning. DXML explores the label correlation by the label graph to alleviate the long tail problem in MLTC. AttentionXML tries to capture the important
parts of texts most relevant to each label, and establishes the relationship between labels and documents, so as to help tail labels capture sufficient semantic information. DXML and AttentionXML can get the satisfying results.
Secondly, for LEAP and OLTR, they all use the transfer strategy to transfer information from the head labels to help the tail labels. Since they are specially designed for long-tailed multi-class classification and don't consider the characteristics of MLTC, thus they are not good at MLTC task. 
Thirdly, HTTN and LTMXP are both specially designed for long-tailed
MLTC task. Both of them train the documents belonging to the head label and the tail label respectively. LTMXP combines linear model and DNN to train the documents belonging to tail label and head label respectively, which ignores the label correlations between the head label and tail labels. HTTN transfers the meta-knowledge from the data-rich head labels to data-poor tail labels. But it doesn't consider whether the transferred information suits for tail labels. As expected, PIRAN consistently outperforms all baselines on all experimental datasets. It further confirms that the proposed Pairwise Instance Relation Augmentation Network is effective for long-tailed multi-label text classification by doing the data augmentation for tail labels.
\subsection{Significance Analysis }
To prove that PIRAN can significantly improves the multi-label text classification  performance on whole labels. The P-value is known as the probability value. It is defined as the probability of getting a result that is either the same or more extreme than the actual observations. The P-value is known as the level of marginal significance within the hypothesis testing that represents the probability of occurrence of the given event. If the P-value is small, then there is stronger evidence in favour of the alternative hypothesis. We give the statistics comparison (P-Value) on the superior baselines on three datasets in terms on F1-score. All experiment results reported in our paper are averaged from five independent runs, thus we list the five results of the superior baseline and PIRAN on Table 5. AttentionXML, LEAP and HTTN are the strongest baselines on AAPD, RCV1 and EUR-Lex respectively, which are  selected as the comparison algorithms for significance calculation. As we all know, P-value $< 0.05$, the result is statistically significant. As shown in Table 5, the P-Value are 0.00019, 0.00037, 0.00013 on three datasets respectively, thus we can make the conclusion that the improvement of PIRAN is statistically significant. 
\subsection{Tail labels Performance Verification }
\begin{table*}[th]
\begin{center}
\caption{\label{font-table4} significance statistics (P-value) on PIRAN and superior baselines in terms of F1-score on three datasets.}
\begin{tabular}{c|cccccc|c|c|}
\hline
Dataset&Methods&1&2&3&4&5&avg&P-value\\
\hline
\multirow{2}{*}{AAPD}&AttentionXML&68.77& 68.85& 68.91& 68.56& 68.86&68.79&\multirow{2}{*}{0.00019}\\
\cline{2-8}& PIRAN&69.32& 69.89& 70.17& 69.69& 69.98&69.81&\\
\hline
\multirow{2}{*}{RCV1}&LEAP&79.48& 79.64& 79.55& 79.44& 79.49&79.52&\multirow{2}{*}{0.00037}\\
\cline{2-8}&PIRAN&79.75& 79.93& 79.88& 79.73& 79.91&79.84&\\
\hline
\multirow{2}{*}{EUR-Lex}&HTTN&53.56& 53.78& 53.21& 53.29& 52.86&53.34&\multirow{2}{*}{0.00013}\\
\cline{2-8}&PIRAN&54.79& 54.87& 54.36& 54.45& 54.93&54.68&\\
\hline
\end{tabular}
\end{center}
\end{table*}
\begin{table}[tp]
\begin{center}
\caption{\label{font-table4} Comparing PIRAN with superior baselines on tail labels.}
\begin{tabular}{|c|ccc|}
\hline
&\multicolumn{3}{|c|}{avg F1 on tail labels}\\
\hline
&AAPD&RCV1&EUR-Lex\\
\hline
AttentionXML&13.23&12.30&1.09\\
\hline
LEAP&16.19&10.17&0.06\\
\hline
HTTN&23.26&16.23& 1.30\\
\hline
PIRAN& \bf{25.90}& \bf{18.78}& \bf{1.96}\\
\hline
\end{tabular}
\end{center}
\end{table}
In order to further verify the performance of these methods on tail labels, we compare the tail label results of PIRAN with that of the strong competitors in multi-label methods (AttentionXML), long-tailed multi-label methods (LEAP) and multi-label methods (HTTN).  Table \ref{font-table4} reports the average of the F1-scores for all tail labels in three datasets. AttentionXML only alleviates the long-tailed problem by constructing the relation between labels and documents, therefore, the performance on the tail labels is poor. LEAP and HTTN are specially designed for long-tailed task. In order to improve the classification performance of tail labels, they try different explorations. LEAP seeks
to expand the distribution of the tail classes during training, so as to alleviate the distortion of the feature space and propose to augment each instance of the tail classes with certain disturbances in the deep feature space. HTTN transfers the meta-knowledge from the data-rich head labels to data-poor
tail labels. The meta-knowledge is the mapping from few-shot network parameters to many-shot network parameters, which aims to promote the generalizability of tail classifiers. HTTN is designed for multi-label learning, the design of HTTN comprehensively considers the characteristics of multi label data. And LEAP is designed for multi-class task, therefore, the performance on the tail labels is worse than HTTN.  PIRAN transfers the instance correlations from head labels to tail labels for generating new instances to enlarge the intra-classdiversity of tail labels, PIRAN truly solves the problem of the imbalance between head labels and tail labels in MLTC task. The results show that PIRAN performs better than the other three strong baselines, verifying again the effectiveness of PIRAN on addressing the tail classification problem. 

\begin{figure}[htbp]
	\centering
\includegraphics[width=8cm]{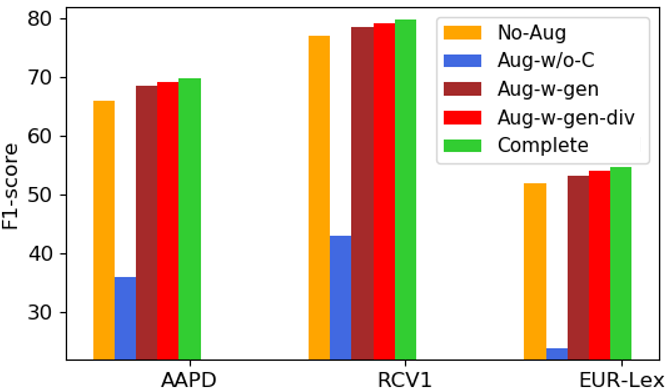}
\caption{Ablation test of PIRAN on three datasets.}
\label{Fig:ablation}
\end{figure}
\begin{figure*}[htp]
\subfigure[AAPD]{
\begin{minipage}[t]{0.33\linewidth}
\centering
\includegraphics[width=5cm]{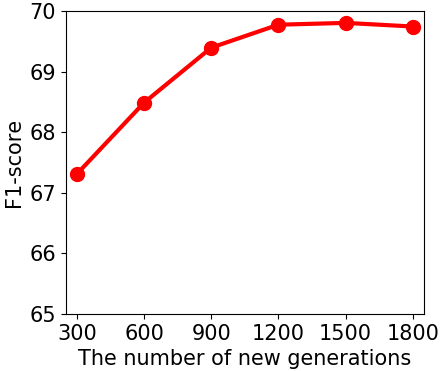}
\end{minipage}%
}%
\subfigure [RCV1 ]{
\begin{minipage}[t]{0.33\linewidth}
\centering
\includegraphics[width=5cm]{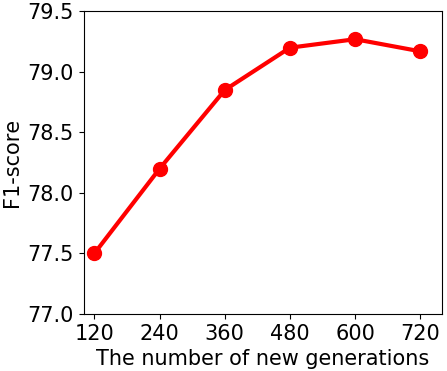}
\end{minipage}%
}%
\subfigure[EUR-Lex]{
\begin{minipage}[t]{0.33\linewidth}
\centering
\includegraphics[width=5cm]{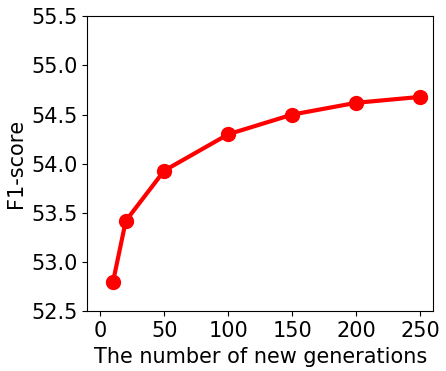}
\end{minipage}%
}%
\caption{Sensitivity of PIRAN to the number of new generations for tail labels ($p$).}
\label{Fig:sens}
\end{figure*}
\begin{figure*}[htbp]
\subfigure[AAPD]{
\begin{minipage}[t]{0.33\linewidth}
\centering
\includegraphics[width=5cm]{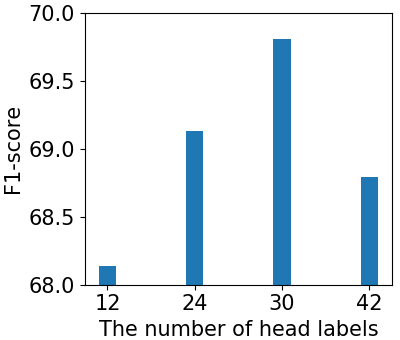}
\end{minipage}%
}%
\subfigure[RCV1]{
\begin{minipage}[t]{0.33\linewidth}
\centering
\includegraphics[width=5cm]{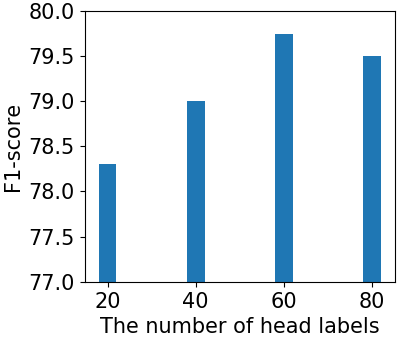}
\end{minipage}%
}%
\subfigure [EUR-Lex ]{
\begin{minipage}[t]{0.33\linewidth}
\centering
\includegraphics[width=5cm]{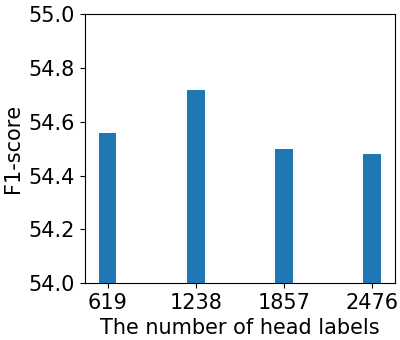}
\end{minipage}%
}%
\caption{Sensitivity of PIRAN to the number of head labels ($l_{head}$).}
\label{Fig:sens4h}
\end{figure*}
\subsection{Ablation Test}
To evaluate the design of different components in PIRAN, we report the  ablation test results in the following setting.
The first one is PIRAN without data augmentation (denoted as \textbf{No-Aug}).  The second one is PIRAN with data augmentation that   however simply generates instances without the constrain of transfer parameter $W$ 
(denoted as \textbf{Aug-w/o-C}).
The third one is PIRAN with data augmentation that generates instances with $W$ but  without  diversity and variance constraint (i.e., just using loss $\mathcal{L}_{gen}$ to train $W$, no  $\mathcal{L}_{div}$ and $\mathcal{L}_{var}$, denoted as \textbf{Aug-w-gen}). The fourth one is PIRAN with data augmentation that generates instances with $\mathcal{L}_{gen}$ and  $\mathcal{L}_{div}$ to ensure the validity of generation instances, denoted as \textbf{Aug-w-gen-div}). The last one is the complete PIRAN (denoted as \textbf{Complete}).
Figure~\ref{Fig:ablation} shows the prediction results on three datasets in terms of F1-score. There are four interesting observations: 
\begin{itemize}
\item It is always preferable to consider the three constraints added in the relation transfer, generation consistency,  generation diversity and distribution consistency, as shown by the superior performance of \textbf{Complete}. It proves that PIRAN can generate high-quality instances for tail labels and use the new instances to improve the performance of tail labels.
\item The result of \textbf{Aug-w/o-C} is the worst, because simple transfer of the correlations without $W$ cannot ensure the effectiveness of the generated instances. It even generates noisy instances to harm the classifier. In other words, learning transfer parameter $W$ is the key to generate high-quality instance. 
\item The result of \textbf{No-Aug} is better than \textbf{Aug-w/o-C}, it shows that generating inaccurate instances will greatly affect the performance. For \textbf{Aug-w/o-C}, it generates a lot of noise data, which really disturbs the learning of tail labels. 
\item  Compared with \textbf{Aug-w-gen},  the model of \textbf{Aug-w-gen-div} can obtain the better result, because the constraint of  $\mathcal{L}_{div}$ avoids redundancy in generating instances. \textbf{Aug-w-gen-div} can effectively enlarge the diversity of tail labels, thus it obtains better result than \textbf{Aug-w-gen} model.
\item The result of \textbf{Complete} is always better than \textbf{Aug-w-gen}, showing the usefulness of $\mathcal{L}_{div}$ and $\mathcal{L}_{dis}$ on improving the classification performance.  
\end{itemize}
Therefore, we can conﬁrm the eﬀectiveness of the three regularizers we designed
in PIRAN. The regularizer of $\mathcal{L}_{gen}$ makes the generations are similiar with tail label's prototype.
Coupling with the regularizer of $\mathcal{L}_{div}$ can eﬀectively enlarge the intra-class diversity of tail labels. Furthermore, $\mathcal{L}_{var}$ further keeps the generated instances for tail labels have a consistent variance with  head labels. 

\subsection{Sensitivity of the Number of New Generations for Tail Labels }
We investigate the performance of PIRAN with different number of new generations for tail labels in AAPD, RCV1 and EUR-Lex dataset. Figure ~\ref{Fig:sens} shows the performance  of PIRAN when the number of generated instances $p$ varies from 300 to 1800 for AAPD, from 120 to 720 for RCV1, from 10 to 250 for EUR-Lex,  and show their inﬂuence on F1-score. For AAPD dataset, when the number of new generations $p$ is changed from 300 to 1200, the classification performance become better. However, the performance improvement reaches a limit when $p$ is as large as 1200. When the number of generated instances $p$ is changed from 1200 to 1800, the classiﬁcation performance is slowly decreased. There is the same trend on the RCV1 dataset,  when $p$ is increased from 120 to 480, F1-score is steadily improved. With the
number of $p$ become larger, the performance of PIRAN is degraded. For EUR-Lex,  as the number of new generations $p$ grows from 10 to 50, the performance dramatically improve, because the number of instance in tail labels mostly only contain 1-3 documents. Therefore, the new generations can effectively improve the classification performance of EUR-Lex at the case of extreme data scarcity. when the number of new generations $p$ increased from 50 to 250,
the classification performance is improved slowly. The expected trend in sensitivity of $p$ confirms that an appropriate number of new generations can improve the classification performance. However, generating too many tail label instances may downgrade  the classification performance due to the introduction of potential noise.

\subsection{Sensitivity of the Number of Head labels }
We vary the number of head labels $l_{head}$, to see how the proposed model is affected by different number of head labels. Figure~\ref{Fig:sens4h} shows the result of different numbers of head labels on AAPD, RCV1 and EUR-Lex evaluation. From the result we can see, for AAPD, from 42 to 30, the classification performance is greatly improved, because when the number of head label is set to 42, some of them are actually tail labels. However, when  it is decreased from 30 to 12, the classification performance is going down due to the few diversity can be transferred. The same trend is shown in RCV1 dataset. when the number of head labels is changed from 20 to 60, the classification performance is greatly improved. However, when the number of head labels increased from 60 to 80, the result is degraded. For EUR-Lex, for the different number of tail labels, the results don't change dramatically.  From 619 to 1238, the classification performance is improved, because the richer instance relationships are brought. However, when  the number of head labels increases from 1238 to 2476,  the performance  of  PIRAN  decreases. The  reason  is  that  many labels  are  really  only  few-shot,  When  they  are  used to learn $Q$, $Q$ may introduce noise. 

In summary, extensive experiment results demonstrate that the proposed PIRAN can achieve competitive performance compared with  baselines. In particular, PIRAN has excellent performance on tail label prediction.


\section{Conclusions and Future Work}
A Pairwise Instance Relation Augmentation Network (PIRAN), in this paper, is proposed for long-tailed multi-label text classification.
It makes use of the instance relations in head labels, and then transfers these relations to tail labels which help tail labels generate new instances in feature space. Two  regularizers (diversity and consistency) are imposed to ensure the generation of high-quality instances in feature space by considering generation consistency, variance consistency and generation diversity. Experiments show that PIRAN consistently outperforms the state-of-the-art baselines and furthermore reveals the performance advantage in predicting long-tailed labels.

In real applications, more precious information can be collected, such as label description, label topology (e.g., hierarchical structure) and etc.
Therefore, it is interesting to extend the current model with such extra information. In the future work, we will transfer the instance correlation from head labels to tail labels by considering the label semantic or label co-occurrence between head and tail labels.

\bibliographystyle{IEEEtran}
\balance
\bibliography{IEEE_TKDE_ICTN}

\end{document}